\documentclass{article}
\usepackage{arxiv}
\usepackage{makeidx}
\usepackage[pdftex]{graphicx}
\usepackage{amsmath,amssymb} 
\usepackage[utf8]{inputenc} 
\usepackage[T1]{fontenc}    
\usepackage{hyperref}       
\usepackage{url}            
\usepackage{booktabs}       
\usepackage{amsfonts}       
\usepackage{nicefrac}       
\usepackage{microtype}      
\usepackage{tikz} 
\usepackage{bm}
\usepackage{color}
\usepackage{subcaption} 
\usepackage{array}
\usepackage{accents}
\newcolumntype{x}[1]{>{\centering\arraybackslash\hspace{0pt}}p{#1}}

\def\secref#1{Sec.~#1}
\def\figref#1{Fig.~#1}

\def\mvec#1{\mathbf{#1}}
\def\ms#1{\mathcal{#1}}
\def\unitCircleLayer{Unit-Circle}
\def\unitCircleLayerShort{U-C}
\def\matcher{Matcher}
\def\irisMatchCNN{IrisMatch-CNN}
\def\norm#1{\left\lVert#1\right\rVert}

\graphicspath{{../pdf/}{../jpeg/}{images/}}
\DeclareGraphicsExtensions{.pdf,.jpeg,.png,.jpg}

\title{Iris Verification with Convolutional Neural Network and Unit-Circle Layer}

	\author{Radim \v Spetl\' ik\thanks{Work performed during an internship at Microsoft Development Center Serbia d.o.o.} \\
  Department of Cybernetics\\
  Czech Technical University\\
  Prague, CZ 120 00 \\
  \texttt{spetrad@cmp.felk.cvut.cz} \\
   \And
 Ivan Razumeni\' c \\
 Microsoft Development Center Serbia\\
 Belgrade, 11070 Serbia\\
 \texttt{irazum@microsoft.com}\\}

\begin{document}
	\maketitle
	\begin{abstract}
		We propose a novel convolutional neural network to verify a~match between two images of the human iris. The network is trained end-to-end and validated on three publicly available datasets yielding state-of-the-art results against four baseline methods. The network performs better by a $10\%$ margin to the state-of-the-art method on the CASIA.v4 dataset. In the network, we use a novel ``\unitCircleLayer{}'' layer which replaces the Gabor-filtering step in a common iris-verification pipeline. We show that the layer improves the performance of the model up to $15\%$ on previously-unseen data.
	\end{abstract}

	\section{Introduction}

Iris verification is a biometric technique used for human identification. Given a~pair of images of human irises, the task is to decide whether the irises match. Iris verification is applied widely, e.g., in border control, citizen authentication, or in forensics \cite{zhao_towards_2017}.

\begin{figure}[t]
    \bgroup
    \begin{tabular}{x{5.5em} x{8em} x{5em} x{5em} x{12em}}
         (a) detection & (b) normalization & (c) feature extraction & (d) mask extraction & (e) matching  \\
    \end{tabular}\\
    \egroup
    \centering
    \includegraphics[width=\columnwidth]{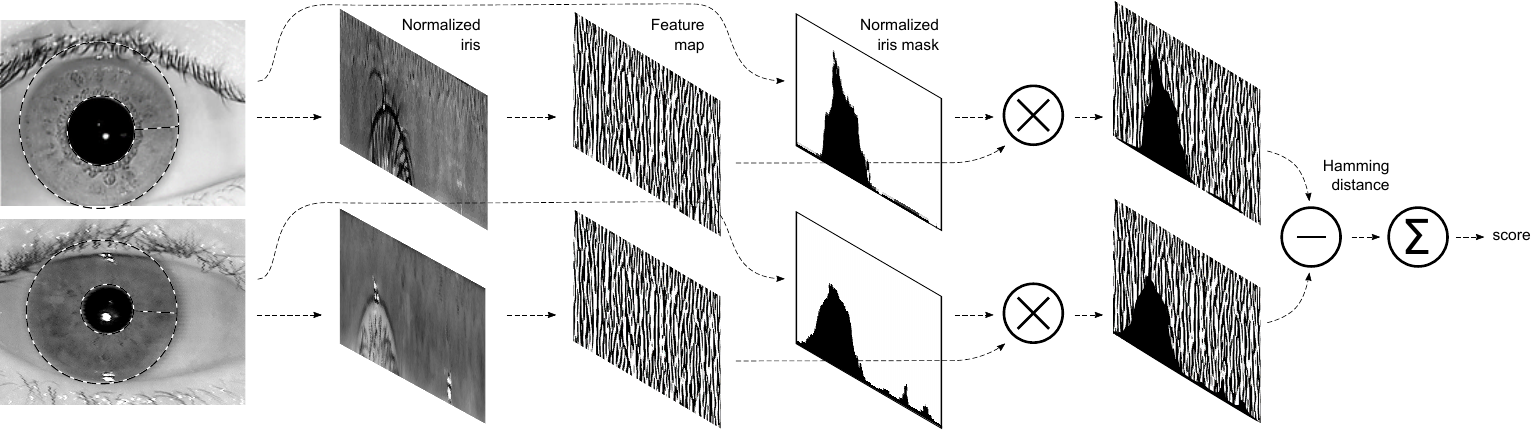}
    \caption{Common iris-verification pipeline. An iris is (a) detected, (b) normalized and (d) its mask is found, (c) the features are extracted. Two irises match (e) if the Hamming distance of their bitcodes multipled by their masks (score) is lower than a given threshold $t$.}
    \label{fig:old-pipeline}
\end{figure}

Common iris verification pipeline has three steps -- iris detection, feature extraction, and matching (see \figref{\ref{fig:old-pipeline}}, interested reader is referred, e.g., to \cite{daugman_how_2004}). First, an iris is found and normalized. Second, the normalized iris is typically convolved with Gabor filters and converted into a ``bitcode'', i.e.\ a matrix of binary numbers. Third, two bitcodes are compared. The bitcodes match if their Hamming distance is smaller than a given threshold. 

\begin{figure}[t]
    \centering
    \includegraphics[width=\columnwidth]{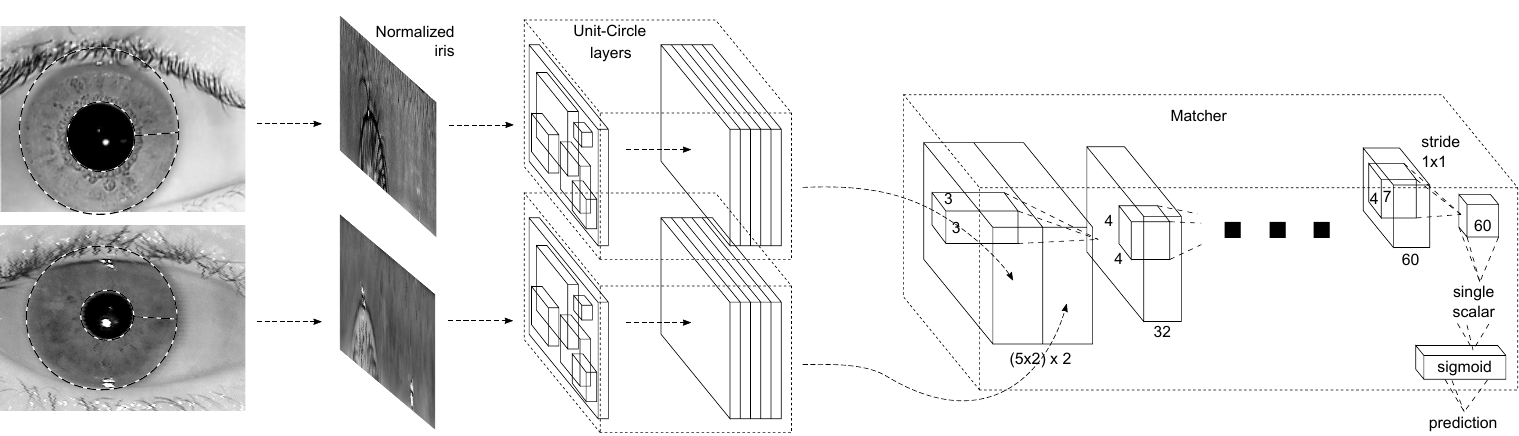}
    \caption{Iris verification with \irisMatchCNN{}. Two irises are detected and normalized. The normalized irises are fed into the \unitCircleLayer{} (\unitCircleLayerShort{}) layers. The responses from the \unitCircleLayerShort{} layers are concatenated and fed into the \matcher{} convolutional network. A single scalar is produced -- the probability of a match. Two irises match if the probability is greater than a given threshold. Compare with a common iris verification pipeline in \figref{\ref{fig:old-pipeline}}.}
    \label{fig:new-pipeline}
\end{figure}

Feature extraction and matching are highly data-dependent in a common iris verification pipeline and therefore require parameter-tuning. Since the task is not convex, an exhaustive search for parameters is performed. In this paper, we propose a method which replaces the feature extraction and matching part of the iris verification pipeline with a single fully convolutional neural network and a single learning rule -- the backward propagation of errors or backpropagation. The network is trained end-to-end using the binary cross-entropy loss function. The input of the network is a pair of normalized irises, the output is a single number which is interpreted as a posterior probability of a match (see \figref{\ref{fig:new-pipeline}}).

So far, convolutional neural networks were used in iris verification for better feature encoding. To encode the features, standard blocks of convolutions, max-pooling, and batch normalization layers were used. We introduce a novel ``\unitCircleLayer{} layer'' that replaces the feature extraction step in a common iris verification pipeline and is learned optimally by backpropagation.

The contributions of this paper are the following: 
(i) we propose a novel method of iris verification that replaces feature extraction and matching steps of a commonly used iris verification pipeline. We replace it with a single convolutional neural network (\irisMatchCNN{}) trained end-to-end that is robust to changes in the iris image acquisition setup, (ii) as opposed to the metric-learning iris verification, we compare two images of irises directly and learn the network with the binary cross-entropy loss, (iii) we evaluate the method on three public datasets against four methods achieving state-of-the-art results.



\section{Related work}

To the best of our knowledge, there is only one work \cite{liu_deepiris:_2016} in which a convolutional neural network (CNN) extracts the features and performs the iris verification at the same time. However, the method is designed to verify a match only between a pair of heterogeneous irises, i.e.\ irises from different sources.

Commonly, researchers in the iris verification domain use a CNN to better encode the iris features. In the following text, we present the methods that use a~CNN at some point in the iris verification pipeline.

\paragraph{CNN as the feature extractor} We start with the methods that use a CNN as the feature extraction tool.

\cite{nguyen_iris_2018} use a pre-trained CNN network to produce a feature vector used for verification. The verification is performed with a support vector machine.

In \cite{zhang_deep_2018}, a deep CNN generates a compact representation of iris and periocular regions. The input of the network is a normalized iris image, the output is a $256$-dimensional feature vector. Cosine similarity, $\ell^1$ norm, $\ell^2$ norm, and covariance measures are used to match two feature vectors. 

In our experiments, we follow the methodology presented in \cite{zhao_towards_2017}. The authors  propose a deep learning framework composed of a CNN that generates iris descriptors and a sub-network that provides a mask identifying iris regions meaningful for matching. The network is trained using a specially designed \emph{Extended Triplet Loss} that incorporates bit-shifting and non-iris masking. The input of the network is a normalized iris image. The output is a feature map that is, together with the mask, used to perform the matching. Matching is done by computing the Hamming distance of two binarized feature maps, taking into account their masks. Experiments on four publicly available databases are presented in which the introduced method outperforms four iris recognition approaches.

Another approach that uses CNN to decode features is presented in \cite{tang_deep_2017}. The learning of the network is formulated as a classification problem. The input is a normalized iris image and output is a $C$-dimensional softmax layer where each class corresponds to a set of irises of a particular person. After the training is finished, the fifth convolutional layer is used as a feature map. To improve robustness, custom ordinal measure is computed that produces a binary vector which is used to perform the matching.

\cite{gangwar_deepirisnet:_2016} present a \emph{deep} CNN that encodes the iris into a $4096$-dimensional feature vector. The learning is stated as a classification problem where each class corresponds to a set of irises of a single person. After the training, the output of the second last fully connected layer is used to compute a similarity score -- the Euclidean distance. The input of the network is a gray-scale iris image normalized to polar coordinates.

\paragraph{CNN used differently than the feature extractor} We follow with three works that use the CNN in another way than just an extractor of the features.

Authors of \cite{liu_deepiris:_2016} design a CNN to verify the relationship between two heterogeneous iris images. A ``pairwise filter'' layer is introduced to extract features from a pair of normalized irises from different sources. For a single pair of irises, six input pairs are generated, explicitly encoding iris rotations and ordering of pairs. The output of the network is a similarity score -- $1$ if the two normalized irises belong to the same identity, $0$ otherwise. Experiments only for irises from heterogeneous sources are presented. It is not clear which loss function is used.

A CNN in \cite{proenca_irina:_2017} distinguishes between corresponding / noncorresponding patches on a normalized iris image. The output of the network is a single scalar -- a probability that the patches correspond. The output of the CNN serves as an input to a Markov random field used to infer a deformation model between a pair of iris images. Given the deformation parameters, the histogram of magnitudes and phase angles are computed. Classification is done with a binary classifier.

A pre-trained ResNet18 in \cite{menon_iris_2018} verifies if two irises match. Despite significant efforts, we were not able to fully comprehend the details of the method. To be more specific: (a) it is not clear what are the inputs of the network, (b) it is not clear why the output of the modified ResNet18 is fed into two perceptrons -- one for the positive class, the other for negative -- and not to a single perceptron, (c) it is not clear, how the outputs of the two perceptrons are used in the final decision.

\section{Method}

\begin{figure}[t]
    \centering
    \includegraphics[width=\columnwidth]{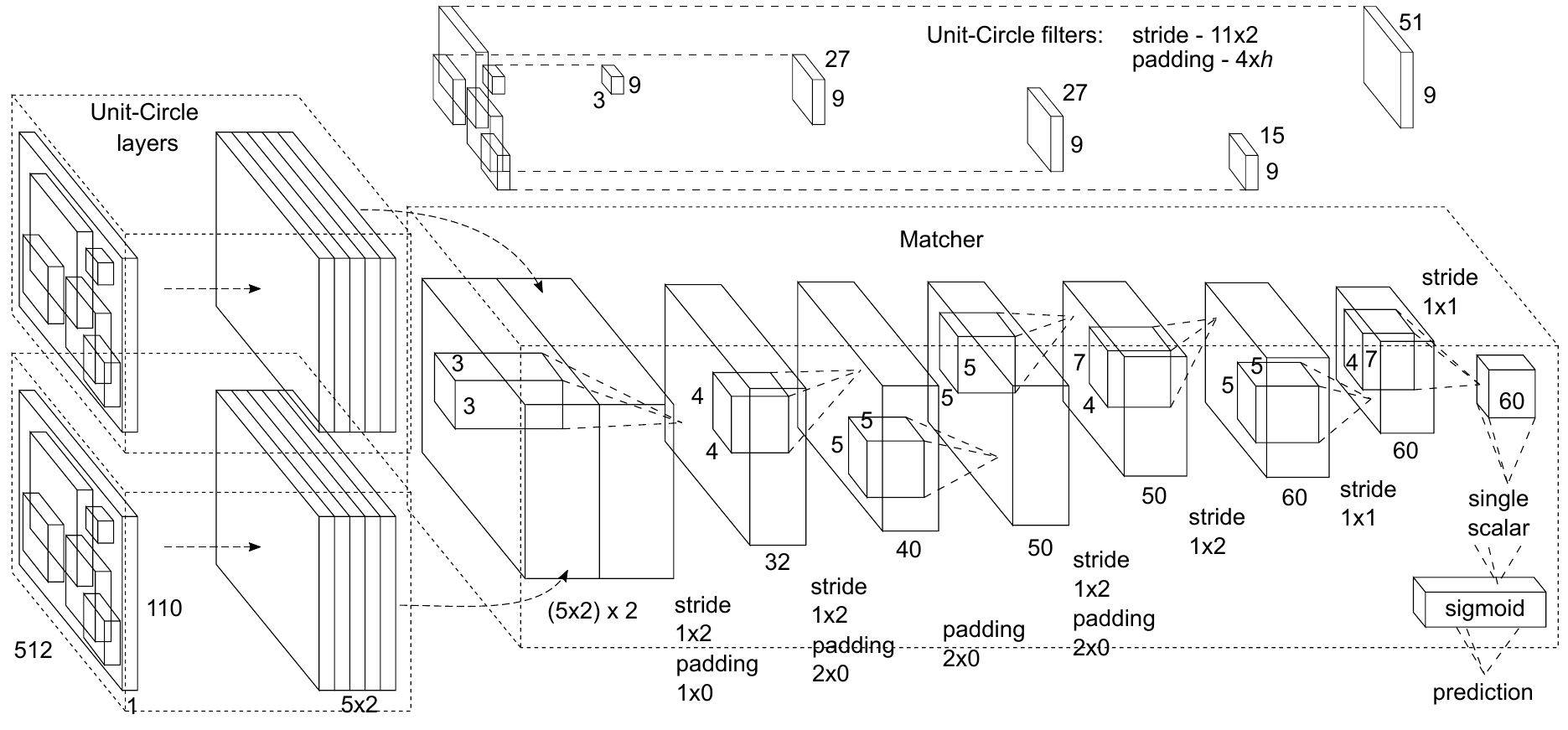}
    \caption{Architecture of the \irisMatchCNN{}. Two normalized-iris images are fed into the \unitCircleLayer{} (\unitCircleLayerShort{}) layers. The responses from the \unitCircleLayerShort{} layers are concatenated and fed into the \matcher{} convolutional network. A~single scalar is produced -- the probability of a match. The variable $h$ stands for an integer part of half the width of the corresponding filter.}
    \label{fig:iris-cnn-architecture}
\end{figure}

We propose a convolutional neural network (CNN) to verify a match of two normalized iris images (see \figref{\ref{fig:iris-cnn-architecture}}). The input of the network is a pair of normalized iris images. The output is a single scalar interpreted as the posterior probability of the match.

The verification has two parts. First, the features are extracted with a novel ``\unitCircleLayer{} layer''. Second, the features are concatenated and fed into the ``\matcher{}'' -- a fully convolutional neural network which outputs a single scalar, the probability of the match. When used together, the \unitCircleLayer{} layer and \matcher{} CNN creates a network architecture to which we refer as to the \irisMatchCNN{}.

Let $\ms{T} = \{(\mvec{x}^j_1,\dots,\mvec{x}^j_{N_j}) \in \ms{X}^{N_j} \mid j = 1, \dots, l\}$ be the training set that contains $l$ tuples of normalized-iris images $\mvec{x} \in \ms{X}$. Each tuple contains $N_j$ images of the same iris. Symbol $\ms{X}$ denotes the set of all input iris images.

\subsubsection{\unitCircleLayer{} layer}
Let $c(\mvec{x}_k; \bm{\phi})$ be the output of a standard convolutional layer with a single input channel and two output channels for the $k$-th normalized iris image, where $\bm{\phi}$ is a~concatenation of the parameters of the filter. We define the output of the \emph{\unitCircleLayer{} layer} $\accentset{\circ}{c}(\mvec{x}_k; \bm{\phi})$ on the $i$-th row and $j$-th column as 
\begin{equation}
    \accentset{\circ}{c}_{(i,j)}(\mvec{x}_k; \bm{\phi}) = \frac{c_{(i,j)}(\mvec{x}_k; \bm{\phi})}{\norm{c_{(i,j)}(\mvec{x}_k; \bm{\phi})}_2}.
\end{equation}
In other words, the output of the \unitCircleLayer{} layer (\unitCircleLayerShort{} layer) is the output of a standard convolutional layer that is normalized along the output channel dimension -- the convolutional layer must have one input channel and two output channels. After the normalization, each pixel in the two-dimensional output of the \unitCircleLayerShort{} lies on the unit cirle.

When multiple \unitCircleLayerShort{} layers are used, we define the concatenation of their responses   $\accentset{\bullet}{c}(\mvec{x}_k; \bm{\Phi}) =  (\accentset{\circ}{c}^1(\mvec{x}_k; \bm{\phi}^1),\dots,\accentset{\circ}{c}^F(\mvec{x}_k; \bm{\phi}^F))$, where $F$ is the number of \unitCircleLayerShort{} layers, $\accentset{\circ}{c}^F(\mvec{x}_k; \bm{\phi}^F)$ is the response of the $F$-th \unitCircleLayerShort{} filter, $\bm{\Phi}$ is a concatenation of all parameters of the $F$ filters.
In \irisMatchCNN{}, five \unitCircleLayerShort{} layers are used, i.e.~we get five pairs of responses or $10$ output channels for each normalized image of iris.

We follow a custom padding strategy. In the vertical direction, we pad by zeroes. Since the normalized image is stored in the polar coordinates, in the horizontal direction and left side of the image we: (i) compute $h$ -- integer part of half the width of the filter, (ii) create a copy of the normalized iris by selecting $h$~pixels from the right side of the image, (iii) append the copy to the left side of the image. We repeat for the right side.

\subsubsection{\matcher{}} The \matcher{} is a~fully convolutional neural network that produces a~single scalar -- the probability that two irises match. Let $g\left(\overline{\mvec{x}}_{q,r}; \bm{\Psi}\right)$ be the output of the \matcher{} CNN for the pair of $q$-th and $r$-th normalized-iris images and $\bm{\Psi}$ a concatenation of all convolutional filter parameters. The input of the \matcher{} CNN is 
$\overline{\mvec{x}}_{q,r} = (\accentset{\bullet}{c}(\mvec{x}_q; \bm{\Phi}),\accentset{\bullet}{c}(\mvec{x}_r; \bm{\Phi}))$, where $\accentset{\bullet}{c}(\mvec{x}_q; \bm{\Phi})$ is the output of \emph{all} \unitCircleLayerShort{} layers for the normalized iris $\mvec{x}_q$ and $\bm{\Phi}$ is a concatenation of the parameters of filters of all \unitCircleLayerShort{} layers.

In other words, the input of the network is created as follows. A normalized iris is fed into the \unitCircleLayerShort{} layers. The output of the \unitCircleLayerShort{} layers is concatenated. The same procedure is repeated for the second normalized iris. Finally, the two sets of responses are concatenated creating the input of the \matcher{} network. Note that the normalized-iris images are fed through the same set of \unitCircleLayerShort{} layers.

\subsubsection{Learning} The binary cross-entropy is used as the objective function. If two irises match, the desired prediction is $1.0$, $0.0$ otherwise. In all experiments, the \matcher{} network was trained first - the weights of the \unitCircleLayerShort{} layers were initialized randomly and fixed. After approx.\ $100$ epochs, we started training the weights of the whole \irisMatchCNN{}. We applied this scheme to speed up the training -- if the whole network was trained from the beginning, the network converged approx.\ $10$ times slower or did not converge at all. The training data are heavily imbalanced towards the negative class (up to $216:1$ ratio). We manually balanced the classes by randomly selecting the $N_p$ negative examples, where $N_p$ is the number of positive examples. We repeated the sub-sampling of the negative class in each epoch.

\paragraph{Technical details} In the \matcher{} CNN, standard blocks of convolutions and Exponential Linear Unit \cite{clevert_fast_2015} activation functions were used. Also, batch normalization and dropout was applied. We trained the network in the \emph{PyTorch 1.0} library with the Adam optimizer and the learning rate set to $0.01$. The set of all input normalized-iris images $\ms{X} = \mathbb{R}^{110 \times 512}$. The output of \emph{all} \unitCircleLayerShort{} layers for the normalized-iris image  $\accentset{\bullet}{c}(\mvec{x}; \bm{\Phi}) = (\accentset{\circ}{c}^1(\mvec{x}; \bm{\phi}_1), \dots, \accentset{\circ}{c}^5(\mvec{x}; \bm{\phi}_5))$. Stride and padding are the same in all five layers.

The training set was split to the training and validation subset with the ratio of $10:1$. \irisMatchCNN{} has $416.930$ parameters in total (compare with approx.\ $12 \cdot 10^6$ of \cite{menon_iris_2018}).

\begin{figure}[t]
  \begin{subfigure}[b]{0.49\textwidth}
    \centering
    \begin{tabular}{c c c}
         \includegraphics[width=6em]{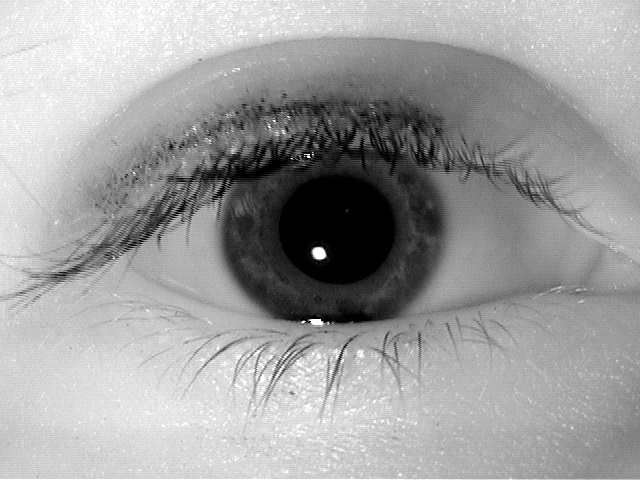} & \includegraphics[width=6em]{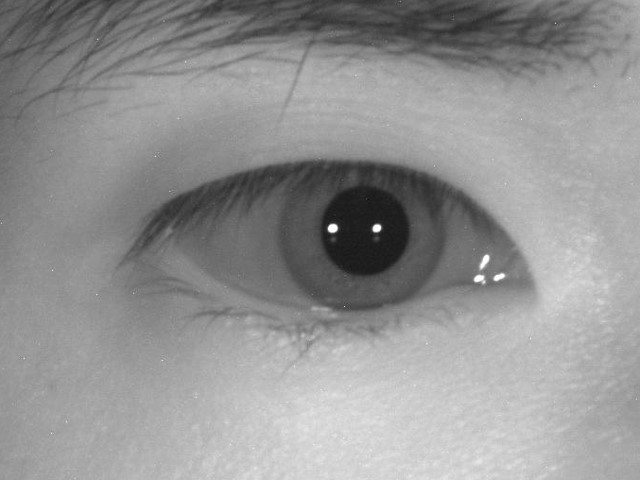} & \includegraphics[width=6em]{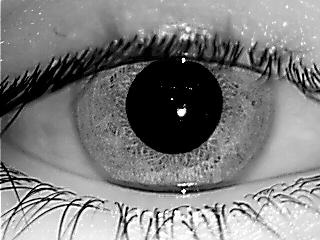} \\
         \includegraphics[width=6em]{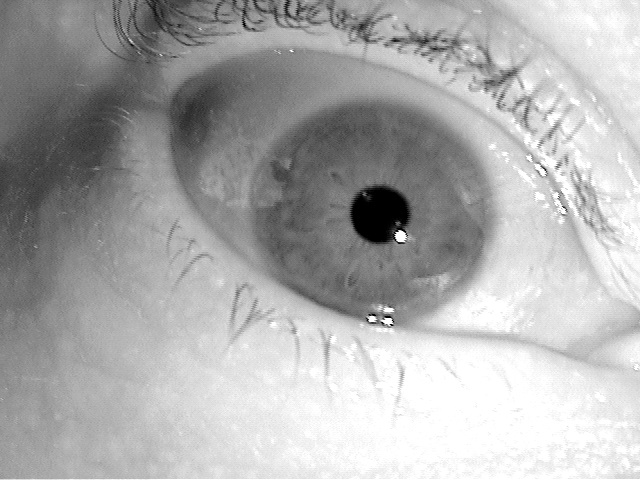} & \includegraphics[width=6em]{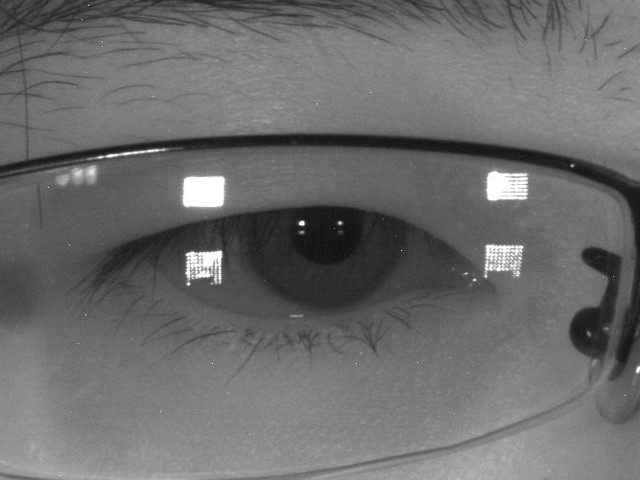} & \includegraphics[width=6em]{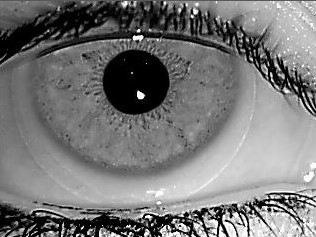} \\
        ND-IRIS-0405 & CASIA V4 & IITD
    \end{tabular}
    \caption{Sample images from ND-IRIS-0405, CASIA v4, and IITD database. Top: dilated pupil, bottom: constricted pupil.}
    \label{fig:dataset-images}
  \end{subfigure}
  \enskip
  \begin{subfigure}[b]{0.47\textwidth}
  \centering
    \begin{tabular}{l r l}
    \toprule
    training & authentic /& imposter \\
    \midrule
    ND-IRIS-0405 & $2,655,213$ /& $563,657,472$ \\
    CASIA v4 & $20,068$ /& $2,619,167$\\
    IITD & $1,410$ /& $277,221$  \\
    \midrule
    testing & authentic /& imposter \\
    \midrule
    ND-IRIS-0405 & $14,791$ /& $5,743,130$ \\
    CASIA v4 & $20,702$ /& $2,969,533$\\
    IITD & $2,240$ /& $624,400$  \\
    \bottomrule
    \end{tabular}
    \caption{Dataset statistics. The number of positive (authentic) and negative (imposter) pairs in the training and testing sets.}
    \label{fig:dataset-statistics}
  \end{subfigure}
  \caption{Overview of the datasets used in the experiments.}
  \label{fig:datasets}
\end{figure}

\section{Experiments}

The quality of iris detection and segmentation has a dramatic effect on the performance of iris recognition pipeline \cite{li_efficient_2019}. Since different authors use different iris segmentation methods, reproducibility of the results reported in iris-verification papers is usually low. Therefore, we follow the methodology of \cite{zhao_towards_2017} -- the authors made their codes public along with the segmentations.

In the experiments, we evaluate the methods with the True Accept Rate (TAR) for a given False Accept Rate (FAR). FAR is a fraction of non-matching pairs classified as matches, TAR is a fraction of matching pairs of iris images classified as non-matches. 

\subsubsection{Datasets}

As discussed earlier in this section, we follow the evaluation procedure introduced in \cite{zhao_towards_2017}. However, we were not able to retrieve the ``WVU Non-ideal Iris Database - Release 1'' since it is currently available only to the residents of the United States. Therefore, we present the results on three datasets -- ND-IRIS-0405, CASIA v4, and IITD. See \figref{\ref{fig:datasets}} for the number of samples in the training and testing subsets and for sample images. In case of all datasets, the iris segmentations provided in the scripts of \cite{zhao_towards_2017} were used to extract the iris in the testing sets. For the training sets,  the irises were segmented with a method introduced in \cite{zhao_accurate_2015}. The models with the highest GAR on the validation subset were selected for the evaluation on the test set.

\paragraph{ND-IRIS-0405} The ND-IRIS-0405 Iris Image Dataset (ICE 2006) \cite{bowyer_nd-iris-0405_nodate} contains $64,980$ iris samples from $356$ subjects. The training set for this database was composed of from the left eye images from all subjects and the test set from the first 10 right eye images from all subjects.

\paragraph{CASIA Iris Image Database V4 - distance} The ``distance'' subset of the CASIA dataset \cite{noauthor_casia.v4_nodate} contains $2,446$ samples from $142$ subjects. The ``distance'' subset is composed of images of the upper part of a face -- each image contains both eyes. The authors of \cite{zhao_towards_2017} provide segmentation of eyes for the subjects in the testing set. For the training set, the eyes were localized with the \emph{IntraFace} facial landmarks detector~\cite{xiong_supervised_2013} using the facial landmarks near the eyes. The training set contains only the right eye images, the test set includes only the left eye images.

\paragraph{IITD Iris Database} The IITD database \cite{noauthor_iit_nodate} includes $2,240$ image samples from $224$ subjects. There are only the right eye images in the training set. In the test set, only the first five left eye images were used.

\begin{figure}[t]
    \centering
    \includegraphics[width=\textwidth]{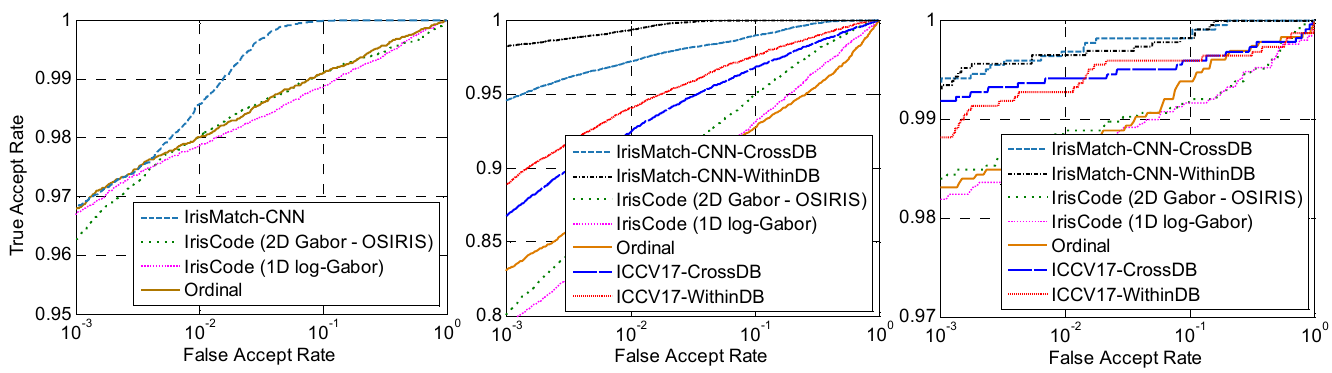}
    \begin{tabular}{x{12em} x{12em} x{12em}}
        (i) ND-IRIS-0405 & (ii) CASIA.v4-distance & (iii) IITD \\
    \end{tabular}
    \caption{Comparative study on three public datasets against four state-of-the-art methods. The proposed \irisMatchCNN{} method yields the best results in the case of all three databases. Results as reported in \cite{zhao_towards_2017}. Note that we were not able to reproduce the results of \cite{zhao_towards_2017} on the ND-IRIS-0405 database. We, therefore, exclude the ICCV17 method from the comparison in case of this database.}
    \label{fig:comparative-study}
\end{figure}

\subsection{Comparative study}

In this experiment, a comparative study on three public datasets against four state-of-the-art methods is presented. 

First, we shortly describe the methods. The most widely deployed iris feature descriptor is the Gabor-filter-based \emph{IrisCode} \cite{daugman_how_2004}. It is a highly competitive method suitable for a performance benchmark \cite{zhao_towards_2017}. A popular public implementation of \emph{IrisCode} is an open source tool for iris recognition \emph{OSIRIS v4.1} \cite{othman_osiris:_2016}. It uses a~band of tunable 2D Gabor filters encoding the iris features at different scales. Another \emph{IrisCode} method \cite{masek_recognition_2003} uses 1D log-Gabor filter(s) to extract the features. \emph{Ordinal} is an approach checking the consistency of Ordinal measures in irises \cite{sun_ordinal_2009}. 

The \emph{ICCV17} and \irisMatchCNN{} methods are presented in two configurations. ``CrossDB`` means that the model was trained only on the training set of the ND-IRIS-0405 database and ``WithinDB'' means that the model was also fine-tuned on the training set of the target database.

We present the results reported in \cite{zhao_towards_2017} -- the results were reproduced with the scripts provided by the authors. However, despite significant efforts, we were not able to reproduce the results in case of ND-IRIS-0405 database. We therefore exclude the ICCV17 method from the comparison in case of this database. Note that in all experiments presented in \cite{zhao_towards_2017}, the vertical resolution of the normalized iris image is $64$ pixels. The \irisMatchCNN{} was developed with the input vertical resolution of $110$ pixels. Therefore, we resize to the required vertical resolution using the linear interpolation. All methods were extensively tuned on the target databases to ensure a fair comparison -- the details are provided in \cite{zhao_towards_2017}.

Taking look at  \figref{\ref{fig:comparative-study}},  \irisMatchCNN{{} yields the best results in case of all three databases. We see that, compared to the other methods, the performance of  \irisMatchCNN{} shows a different trend -- TAR tends to be higher for a wider range of FAR, which is especially visible in case of the ND-IRIS-0405 database. We believe, that this tendency is caused by the binary cross-entropy objective function. 

Let us examine both ``CrossDB'' and ``WithinDB'' setup in \figref{\ref{fig:comparative-study}} now. The results of \irisMatchCNN{} on the IITD database do not differ much between these two settings. A difference of approx.\ $3$ per cent in favour of ``WithinDB'' setup is visible in case of the CASIA database. We conclude that \irisMatchCNN{} generalizes well between different databases, i.e.\ the method is robust to changes in the iris image acquisition setup.

\subsection{Effect of \unitCircleLayer{} layers on performance}

\begin{figure}[t]
    \centering
    \includegraphics[width=\columnwidth]{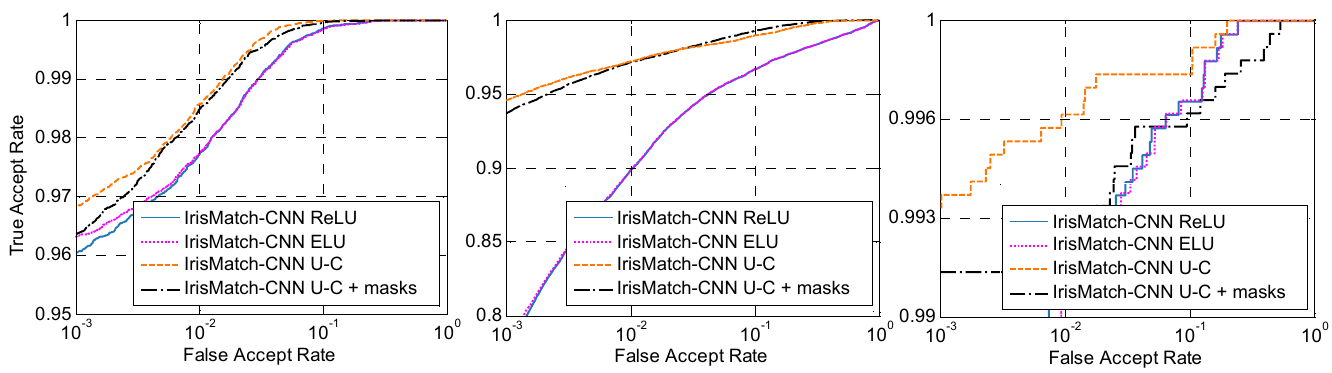}
    \begin{tabular}{x{12em} x{12em} x{12em}}
        (i) ND-IRIS-0405 & (ii) CASIA.v4-distance & (iii) IITD\\
    \end{tabular}
    \caption{Effect of the \unitCircleLayer{} layers on the performance of \irisMatchCNN{}. The $\ell_2$ normalization was replaced by the ReLU and ELU non-linearities and the model was learnt for $1000$ epochs. In ``masks'' experiment, iris masks were concatenated to outputs of the \unitCircleLayer{} layers.}
    \label{fig:cr-layers-study}
\end{figure}

We developed the \unitCircleLayer{}  (\unitCircleLayerShort{}) layer as a replacement of the Gabor filtering step in the iris recognition pipeline. We interpret the outputs of the \unitCircleLayerShort{} layer as responses lying on the unit circle in a two-dimensional plane. In the following experiment, we replaced the $\ell^2$ normalization in the \unitCircleLayerShort{} layer by two non-linearities -- by the Rectified Linear Unit, or ReLU, and the Exponential Linear Unit \cite{clevert_fast_2015}, or ELU. We trained the \irisMatchCNN{} network with each non-linearity on the training set of the ND-IRIS-0405 database for $1,000$ epochs, selecting the model with the best validation performance.

As seen in \figref{\ref{fig:cr-layers-study}} in case of all three databases, the TAR is higher for the network in which the \unitCircleLayerShort{} layers are used. From the plots (ii) and (iii), we conclude that the \unitCircleLayerShort{} layers improve generalization on unseen data.

\subsection{Effect of iris masks on performance}

The input of \irisMatchCNN{} is a pair of normalized irises. In this experiment, we also included the masks estimated by the ICCV17 mask sub-network so that the input of the \irisMatchCNN{} network is a pair of normalized irises with their masks. The normalized-iris mask is shown in \figref{\ref{fig:old-pipeline}}. In a common iris-verification pipeline, it marks the areas not suitable for matching -- e.g., eyelids, sclera, or reflections.

The results are shown in \figref{\ref{fig:cr-layers-study}}. There is no significant improvement when also the masks are included. Therefore, we conclude that the \irisMatchCNN{} network is capable of determining the ``good areas to match'' by itself.

\subsection{Effect of iris-segmentation method on \irisMatchCNN{} performance}
\label{sec:segmentation}

In this experiment, the task was to examine the robustness of \irisMatchCNN{} against the change of the iris segmentation method. In our datasets, we segment the iris with the total variation method \cite{zhao_accurate_2015}. In this experiment, we employed a publicly available iris-verification software \emph{OSIRIS v4.1} that uses the Viterbi method \cite{sutra_viterbi_2012}. We created two modified test subsets from the ND-IRIS-0405 and CASIA.v4 databases (see \figref{\ref{tab:segmentation-dataset}} for statistics). We followed the left/right eye splits in the testing datasets. However, we used all images that were successfully extracted by both the methods. The presented results were retrieved by the  \irisMatchCNN{} network that was trained on the standard training subset of the ND-IRIS-0405 database.

The first thing that needs a comment is a much higher performance visible in \figref{\ref{fig:segmentation-results}} in case of ND-IRIS-0405. The test set used in other experiments contains only the first $10$ right eye images from all subjects. In this experiment, we did not follow this limitation. Instead, we used all iris images that were successfully segmented by both the methods. This condition resulted in an ``easy-to-verify'' set of normalized irises. However, the first conclusion of this experiment -- for the database on which the \irisMatchCNN{} was trained the switch between the total variation method and the Viterbi method makes no difference in performance.

However, the results on the CASIA.v4-distance database give us a different view. The total variation method gives better TAR by $10\%$ for a $10^{-3}$ FAR than the Viterbi method. We conclude, that the \irisMatchCNN{} method is not robust to changing the capture settings (i.e.\ the database) and the segmentation method at the same time.

\begin{figure}[t]
    \centering
    \begin{subfigure}[b]{0.66\textwidth}
        \includegraphics[width=\columnwidth]{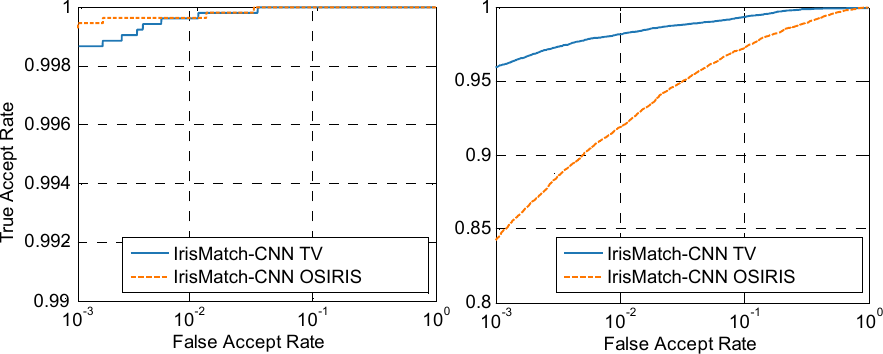}
        \begin{tabular}{x{12em} x{12em}}
             (i) ND-IRIS-0405 & (ii) CASIA.v4-distance \\
        \end{tabular}
        \caption{Iris-segmentation experiment results. Solid blue: irises segmented with the total variation method \cite{zhao_accurate_2015}, orange dashed: irises segmented with the Viterbi method \cite{sutra_viterbi_2012}.}
        \label{fig:segmentation-results}
    \end{subfigure}
    \enskip
    \begin{subfigure}[b]{0.31\textwidth}
        \centering
        \begin{tabular}{l r r}
            \toprule
            authentic & &  \\
            \midrule
            ND-IRIS-0405 &  & $21,886$ \\
            CASIA v4 & & $13,231$\\
            \midrule
            imposter & &  \\
            \midrule
            ND-IRIS-0405 & & $13,505,915$ \\
            CASIA v4 & & $1,711,922$\\
            \bottomrule
        \end{tabular}
        \caption{Statistics of test dataset used in  iris-segmentation experiment.\\\vspace{0.7em}}
        \label{tab:segmentation-dataset}
    \end{subfigure}
    \caption{Effect of iris-segmentation method on \irisMatchCNN{} performance. The network was trained only on the ND-IRIS-0405 training set. The test datasets are described in \secref{\ref{sec:segmentation}} (see also \figref{\ref{tab:segmentation-dataset}}).}
    \label{fig:sementation-experiment}
\end{figure}

Note that we excluded the IITD database from this experiment since the number of authentic pairs in the testing set, which we got by the previously described construction, was less than $1,000$.

\subsection{Heterogeneous iris verification}

In this experiment, we inspect the performance of the \irisMatchCNN{} model in the heterogeneous iris verification. In this type of verification, two images of irises are compared, but each iris is from a different source (as opposed to the previous experiments, in which the pair of irises was always from the same source).

For the purpose of this experiment, we use the ND-CrossSensor2013 database\footnote{Available at \href{https://cvrl.nd.edu/projects/data/\#nd-crosssensor-iris-2013-data-set}{https://cvrl.nd.edu}.}. In this database, each iris is captured with both the LG2200 and LG4000 iris sensors. We follow the experimental protocols introduced by the authors of the database. More specifically, we use the ``SigSets2013-Small-LG4000-LG2200'' protocol that specifies which irises should be compared. In this protocol, there is a~total number of $99,690,130$ comparisons. However, we segmented the normalized irises with the total variation method \cite{zhao_accurate_2015} and we were not able to segment the whole dataset. Therefore, in our experiment, there is a total number of $18,000,588$ comparisons (see \figref{\ref{tab:cross-sensor-dataset}} for numbers of positive class, or authentic, and negative class, or imposter, pairs).

Let us take a look at the results in \figref{\ref{fig:cross-sensor-results}}. In the first experiment, we used the model trained on the ND-IRIS-0405 database training subset (see \figref{\ref{fig:comparative-study}} for results on other datasets). Then, we tried to fine-tune the model on a training subset of the ND-CrossSensor2013 database. Compared to the results reported by the ACSTL Cross-Sensor Comparison Competition Team 2013 \cite{popescu-bodorin_cross-sensor_2018}, our models perform more than $20\%$ worse at the false accept rate of $10^{-3}$. We believe that this is caused by the architecture of the \irisMatchCNN{} -- we use the same set of \unitCircleLayer{} layers for both input  images of irises. In fact, our experiment verifies results of \cite{liu_deepiris:_2016}. The authors design a special ``pairwise bank filter'' to account for differences between the heterogeneous irises, i.e.\ irises from different sources. In our case, the source of the difference between the irises is the capturing device. Compared to the LG4000-based sensors, the LG2200-based sensors produce blurry images commonly with a strong interlacing (see \figref{\ref{fig:cross-sensor-irises}}}). We conclude, that \irisMatchCNN{} is not suitable for heterogeneous iris verification.

\begin{figure}[t]
    \centering
    \begin{subfigure}[b]{0.28\textwidth}
    \def\arraystretch{1}
    \setlength\tabcolsep{1.5em}
        \begin{tabular}{c c}
             \includegraphics[width=0.49\columnwidth]{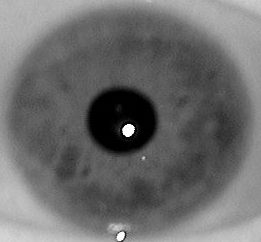} &  \includegraphics[width=0.49\columnwidth]{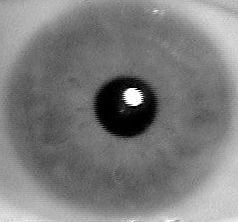} \\
             \includegraphics[width=0.49\columnwidth]{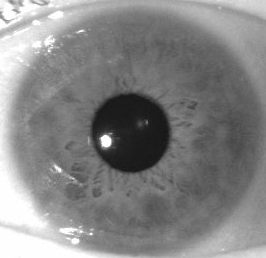} & \includegraphics[width=0.49\columnwidth]{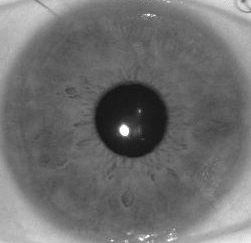} \\
             left eye & right eye\\
        \end{tabular}
        \caption{ND-CrossSensor2013 database. Detail of iris, same subject. Top: LG2200, bottom: LG4000 sensor.}
        \label{fig:cross-sensor-irises}
    \end{subfigure}
    \enskip
    \begin{subfigure}[b]{0.37\textwidth}
        \includegraphics[width=\columnwidth]{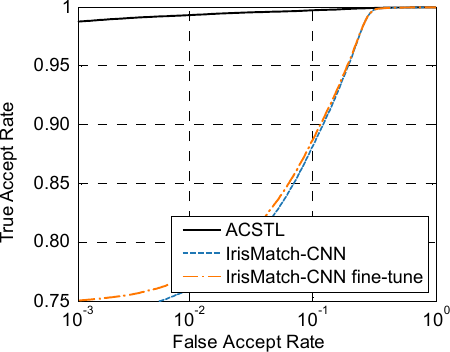}
        \caption{\irisMatchCNN{} evaluated on ND-CrossSensor2013 database (see \figref{\ref{tab:cross-sensor-dataset}}). Solid: ACSTL as reported in \cite{popescu-bodorin_cross-sensor_2018}, dashed: \irisMatchCNN{} trained on ND-IRIS-0405 database, dot dash: fine-tuned.}
        \label{fig:cross-sensor-results}
    \end{subfigure}
    \enskip
    \begin{subfigure}[b]{0.27\textwidth}
        \centering
        \begin{tabular}{l r r}
            \toprule
            \multicolumn{3}{l}{\textbf{ND-CrossSensor2013}} \\
            \midrule
            training & &  \\
            \midrule
            authentic &  & $15,071$ \\
            imposter & & $2,366,456$\\
            \midrule
            testing & &  \\
            \midrule
            authentic & & $41,876$ \\
            imposter & & $17,958,712$\\
            \bottomrule
        \end{tabular}
        \caption{Statistics of the testing and training subsets of ND-CrossSensor2013 database.\\\vspace{0.7em}}
        \label{tab:cross-sensor-dataset}
    \end{subfigure}
    \caption{Heterogeneous iris verification. Evaluated on the ND-CrossSensor2013 database. We used the \irisMatchCNN{} model from the ND-IRIS-0405 comparative study. For the fine-tuning, the same model was trained for $1,000$ epochs and the one yielding the best validation true accept rate for $10^{-3}$ false accept rate was selected.}
\end{figure}

\section{Conclusion}

In this paper, we introduced a novel convolutional neural network architecture \irisMatchCNN{} yielding state-of-the-art results in the iris-verification task on the ND-IRIS-0405, CASIA.v4-distance, and IITD databases. The input of the network is a normalized-iris image, the output is a single scalar interpreted as the probability of a match. A novel \unitCircleLayer{} layer was introduced that improves robustness of the model (i.e.\ the ability of the model to generalize on previously-unseen data), which is verified in experiments. We presented experiments in which a different iris-segmentation method: (a) does not affect the performance when evaluated on previously-seen data (b) decreases the performance otherwise. Lastly, we showed that \irisMatchCNN{} is not suitable for heterogeneous iris verification, i.e.\ for matching two irises when each is from a~different source.

The input of the \irisMatchCNN{} model is a normalized-iris image. Therefore, the performance of the model heavily depends on the iris detection and segmentation methods. To the best of our knowledge, there is no work in which the detection, segmentation, and matching is performed end-to-end. We believe that the next steps in the iris verification domain will incorporate the detection and segmentation into a customized neural-network architecture that will yield excellent results and will be compact at the same time.



\bibliographystyle{splncs03}
\bibliography{references}

\end{document}